\documentclass[letterpaper, 10 pt, conference]{ieeeconf}  % Comment this line out if you need a4paper

\IEEEoverridecommandlockouts                              % This command is only needed if 
\usepackage{cite}
\usepackage{amsmath,amssymb,amsfonts}
\usepackage{multirow}
\usepackage{float}
\usepackage{subfigure}
\usepackage{comment}
\usepackage{algorithm}
\usepackage{algpseudocode}
\usepackage{booktabs} % For prettier tables
\usepackage{array}    % For better column definitions
\usepackage{siunitx}  % For unit and number formatting
\usepackage{graphicx} % For including images
\usepackage[table]{xcolor} % For cell coloring
\usepackage{hyperref}
\hypersetup{
    colorlinks=true,    % 设置链接颜色
    linkcolor=blue,     % 文档内部链接颜色（如\ref{...}和\pageref{...}）
    filecolor=magenta,  % 文件链接颜色
    urlcolor=cyan,      % URL链接颜色
    citecolor=green,    % 参考文献链接颜色
}

\title{\LARGE \bf
LDP: A Local Diffusion Planner for Efficient Robot Navigation and Collision Avoidance
}

\author{Wenhao Yu$^{1}$, Jie Peng$^{2}$, Huanyu Yang$^{2}$, Junrui Zhang$^{1}$, Yifan Duan$^{3}$, Jianmin Ji$^{3, *}$ and Yanyong Zhang$^{3}$%$% <-this % stops a space
\thanks{$^{1}$ Institute of Advanced Technology, University of Science and Technology of China (USTC), Hefei 230026, China
        {\tt\small wenhaoyu@mail.ustc.edu.cn}}%
\thanks{$^{2}$ School of Data Science, USTC, 230026, China}%
\thanks{$^{3}$ School of Computer Science and Technology, USTC, 230026, China}%
\thanks{${*}$ Corresponding author. {\tt\small jianmin@ustc.edu.cn}}
}

\begin{document}

\maketitle
\thispagestyle{empty}
\pagestyle{empty}

%%%%%%%%%%%%%%%%%%%%%%%%%%%%%%%%%%%%%%%%%%%%%%%%%%%%%%%%%%%%%%%%%%%%%%%%%%%%%%%%
\begin{abstract}

The conditional diffusion model has been demonstrated as an efficient tool for learning robot policies, owing to its advancement to accurately model the conditional distribution of policies.
The intricate nature of real-world scenarios, characterized by dynamic obstacles and maze-like structures, underscores the complexity of robot local navigation decision-making as a conditional distribution problem. Nevertheless, leveraging the diffusion model for robot local navigation is not trivial and encounters several under-explored challenges:
(1) \textit{Data Urgency} The complex conditional distribution in local navigation needs training data to include diverse policy in diverse real-world scenarios;
(2) \textit{Myopic Observation} Due to the diversity of the perception scenarios, diffusion decisions based on the local perspective of robots may prove suboptimal for completing the entire task, as they often lack foresight. In certain scenarios requiring detours, the robot may become trapped.
To address these issues, our approach begins with an exploration of a diverse data generation mechanism that encompasses multiple agents exhibiting distinct preferences through target selection informed by integrated global-local insights.
Then, based on this diverse training data, a diffusion agent is obtained, capable of excellent collision avoidance in diverse scenarios.
Subsequently, we augment our \underline{L}ocal \underline{D}iffusion \underline{P}lanner, also known as \texttt{LDP} by incorporating global observations in a lightweight manner.
This enhancement broadens the observational scope of \texttt{LDP}, effectively mitigating the risk of becoming ensnared in local optima and promoting more robust navigational decisions.
Our experimental results demonstrated that the \texttt{LDP} outperforms other baseline algorithms in navigation performance, exhibiting enhanced robustness across diverse scenarios with different policy preferences and superior generalization capabilities for unseen scenarios. Moreover, we highlighted the competitive advantage of the \texttt{LDP} within real-world settings.

\end{abstract}

%%%%%%%%%%%%%%%%%%%%%%%%%%%%%%%%%%%%%%%%%%%%%%%%%%%%%%%%%%%%%%%%%%%%%%%%%%%%%%%%
\section{Introduction}

With the rapid advancement of artificial intelligence and robotics, an increasing number of technologies are being integrated into motion planning for robot collision avoidance~\cite{xiaoMotionPlanningControl2022a}. Many learning-based methods model the planning task as a conditional probability generation problem, where the robot's action sequence is a latent variable with its prior distribution~\cite{cui2022play,chi2023diffusion}. The planning process is accomplished by computing the posterior distribution based on conditions such as robot observations, final rewards, and constraints. 
% All these works can be encompassed within the framework of \textit{``planning as inference''}~\cite{botvinick2012planning, attias2003planning}.

%~\Jie{Logic should be Complex Environment $\to$ (1) More data (more diverse data), (2) Complex environment need border foresight}~\ynote{Done}
% It is widely acknowledged that data plays a crucial role in applying probability theory to solve practical problems, such as in the fields of statistical inference, machine learning, and data science. The task addressed in this paper similarly underscores the critical importance of data.
% One of the key insights guiding our work is that \textit{``if we can capture the data, then the robot can accomplish the task corresponding to the policy data.''} However, this presents considerable challenges. In the task of robot collision avoidance, the diversity of real-world scenarios leads to multimodality in expert policy data distribution. The multimodal nature of expert policy data distribution demands a high expressiveness from the policy model. 
As is well known, the real-world environment for robot navigation is complex, encompassing various scenarios. Designing a specific policy for each scenario would require immense effort and lack the necessary flexibility and scalability. Therefore, an outstanding navigation policy must effectively handle diverse scenarios. Furthermore, the distribution of near-optimal expert policies often varies across different scenarios, highlighting the need for navigation policies capable of addressing diverse scenarios to exhibit a multimodal distribution. Given these requirements, collecting expert data from diverse scenarios to use as model training data and constructing models that better represent multimodal distributions will become crucial. Additionally, the limited local perspective of robots often fails to provide sufficient information for devising policies to tackle diverse scenarios. Robots frequently fall into suboptimal states during navigation tasks. This situation is commonly known as a local minimum problem~\cite{WANG2008625}.

\begin{figure}[t]
    \centering
    \includegraphics[width=0.8\columnwidth]{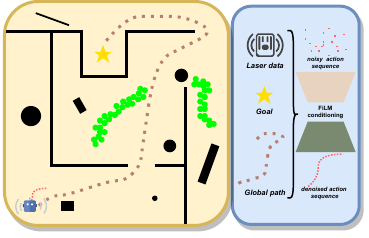}
    \caption{The diagram illustrates the execution of our method. Obstacles are denoted by black circles and rectangles, while the trajectories of pedestrians are represented by green circles. The navigation target is marked by a yellow pentagram, and a brown dashed line delineates the global path from the robot's starting point to its target. This system utilizes laser data, target point information (goal), and the global path to generate local action sequences in a classifier-free guidance approach. 
    }
    \label{fig:first}
\end{figure}

Therefore, to address the challenges mentioned above, we have made two efforts in this paper: (1) we have collected expert policy data with multiple preferences under diverse scenarios and utilize the diffusion model, which has strong distribution modeling capabilities, to construct the policy model. (2) We incorporate global paths as an additional condition to guide the diffusion model, enhancing policies for better navigation through maze-like scenarios and similar environments. Specifically, we have collected expert policy data in three different types of scenarios, i.e., dense static, dynamic pedestrian, and maze-like. For each scenario, we have gathered expert policy data for two preferences, i.e., the original Soft Actor-Critic (SAC) policy~\cite{yu2023pathrl} and the SAC policy guided by global paths. Using the Denoising Diffusion Probabilistic Models (DDPM)~\cite{ho2020denoising} algorithm, based on robot observations (local costmaps, goals, and global paths), we directly denoise and sample from the posterior trajectory distribution step by step, generating the final robot action sequence.

In our experiments, the \texttt{LDP} outperforms other baseline algorithms across various scenarios. It exhibits exceptional learning capabilities when dealing with data from mixed scenarios, displaying remarkable robustness. Moreover, it demonstrates enhanced performance in unseen scenarios, showcasing impressive zero-shot generalization abilities. The addition of the global path as conditional guidance enhances our policy's capacity to comprehend sample distribution, leading to more forward-thinking policy outcomes. Furthermore, our approach leverages the advantages offered by expert policy data from two preferences, resulting in more effective decision-making in diverse navigation scenarios. Ultimately, we deploy our policy algorithm in real-world scenarios and on robotic platforms, thereby showcasing its competitive advantages.

The main contributions of our work are summarized as follows:
\begin{itemize}
    \item The paper introduces \texttt{LDP}, a novel local planning algorithm for robotic collision avoidance, leveraging diffusion processes. 
    % This method fully exploits the potent distribution modeling capability in diffusion models, facilitating the learning of exceptional policies across various scenarios and expert data with differing preferences.
    \item We provide a dataset of expert policy based on 2D laser sensing, which spans expert data across three different types of scenarios and two different preferences.
    \item With global paths serving as additional guiding conditions, the diffusion model can better learn the distribution of expert data and make wiser decisions.
    \item We conducted extensive experiments demonstrating that \texttt{LDP} outperforms other baseline algorithms in terms of superior navigation performance, stronger robustness, and more profound generalization capabilities. Furthermore, we validated the effectiveness of the algorithm by deploying it on physical robots, thus highlighting its practical value.
\end{itemize}

\section{Related Work}

\subsection{Traditional Navigation Approaches}
Traditional navigation systems~\cite{xiaoMotionPlanningControl2022a,peng2021towards} typically adopt a hierarchical paradigm that combines global and local path planning with motion control. These methods can be broadly classified into three categories: search-based planners (e.g., hybrid A*~\cite{Kurzer1057261}, JPS~\cite{harabor2011online}), sampling-based planners (e.g., PRM~\cite{kavraki1996probabilistic} and RRT~\cite{lavalle2001rapidly}), and optimization-based planners (e.g., TEB~\cite{rosmann2012trajectory}). Despite their prevalent usage, a significant amount of effort is needed to tune parameters for these methods to adapt to a wide range of scenarios.

\subsection{Learning-Driven Navigation Approaches}
Imitation learning (IL) is a process that involves learning from examples provided by an expert, typically in the form of decision-making data from human operators. IL has a broad and profound influence in areas such as robotic navigation~\cite{yu2023pathrl,yao2021crowd,qiu2022learning}, manipulation~\cite{chi2023diffusion,chi2024universal}, and autonomous driving~\cite{pomerleau1988alvinn,you2023p,bojarskiEndEndLearning2016}. Depending on the structure of the policy model construction, these methods can be bifurcated into two categories:

\subsubsection{Explicit Policy}
These methods learn the mapping from observations to actions directly, guiding the policy learning process via a regression loss. \cite{bojarskiEndEndLearning2016} employed behavior cloning (BC) to train the end-to-end deep convolutional neural network (CNN) for autonomous driving. The network ingests images from a car camera and yields the steering wheel angle for the vehicle. However, these policies often grapple with the effective modeling of multimodal data distributions. They usually map one policy to one scenario, and learning from data across multiple scenarios can lead to catastrophic forgetting. Furthermore, they often face challenges in generalizing to new, unseen scenarios.

\subsubsection{Implicit Policy}
These methods employ energy-based models(EBMs)~\cite{du2019implicit} to represent action distributions. Each action is assigned an energy value, and the action prediction problem is transformed into an optimization problem to find the action with the lowest energy. This implicit design approach can more effectively represent the multimodal distribution of expert actions.

% Our approach also aligns with the design of imitation learning. The entire task is characterized as a \textit{planning as inference} problem. The key difference is that we use a diffusion model to model the gradient of the expert action distribution score function, and iteratively optimize the gradient field via a series of stochastic Langevin dynamics steps. This design ensures the model's capability to represent multimodal distributions and is more advantageous for high-dimensional decision-making and training stability.

\subsection{Diffusion Model for Robotic Decision Planning}
In recent years, a growing number of works have leveraged diffusion models to perform intelligent agent decision planning tasks, including imitation learning~\cite{chi2023diffusion,carvalho2023motion,sridhar2023nomad} and reinforcement learning~\cite{janner2022planning,zhu2023diffusion,he2024diffusion}. Diffuser~\cite{janner2022planning} concatenates state-action sequences of a certain length into a two-dimensional array, employs the original DDPM method for unconditional sampling, and designs a classifier based on rewards, goals, and other information to guide the inference denoising process, thereby ensuring that the generated decision sequences comply with the respective constraints. The diffusion policy~\cite{chi2023diffusion} represents an alternative modeling method that uses the robot's visual observations as conditions and directly guides the generation of action sequences in a classifier-free manner. MPD~\cite{carvalho2023motion} integrates diffusion models and optimization-based methods. With the planning of start and end points and various optimization costs as conditions, it guides the generation of global motion planning in static scenarios. However, \texttt{LDP} accomplishes local motion planning in diverse scenarios (static and dynamic environments). NoMaD~\cite{sridhar2023nomad} harnesses the powerful distribution modeling capability of diffusion models and employs a goal mask to achieve a single policy capable of completing both navigation and exploration tasks. In contrast, \texttt{LDP} evaluates the navigation performance of the policy model under a mixed expert trajectory distribution of multiple scenes and preferences and introduces additional global paths as conditions for guidance.

% Our method is analogous to the MPD and NoMaD methods. However, unlike MPD, \texttt{LDP} accomplishes local motion planning in diverse scenarios (static and dynamic environments). Unlike NoMaD, \texttt{LDP} evaluates the navigation performance of the policy model under a mixed expert trajectory distribution of multiple scenes and preferences and introduces additional global paths as conditions for guidance.

\begin{figure*}[h]
    \centering
    \includegraphics[width=1.0\textwidth]{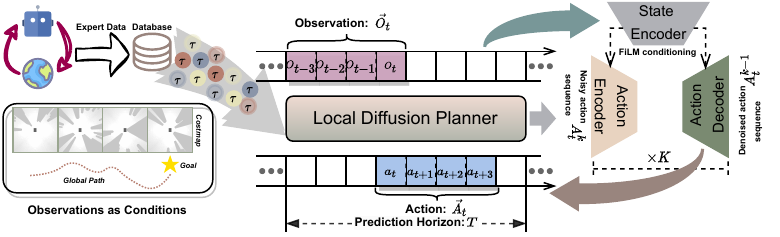}
    \caption{An in-depth depiction of the entire process and the architecture of the local diffusion planner. The circles, marked with different colors and denoted as $\tau$, represent expert data gathered from diverse scenarios, each reflecting various preferences through distinct levels of transparency. At time step $t$, the planner takes in observations $\Vec{O}_t$ from the past $T_o$ steps and predicts the action sequence $\Vec{A}_t$ for the next $T_a$ steps. During the training process, the loss is computed based on DDPM noise prediction over the entire sequence $T$. In the inference process, an action sequence can be generated for every $K$ iterations of denoising.}
    \label{fig:overview}
\end{figure*}

\section{Method}

%\Jie{In this section, we first introduce how we collect training data target on the \textit{Data Urgency} challenge, and then we move on to the details of how to elaborate the Diffusion Model for local planning with the data we collected while aiming the \textit{Myopic Observation} problem.}~\ynote{Done}
In this section, we initially outline our approach to gathering training data focusing on the \textit{Data Urgency} challenge. Subsequently, we delve into the specifics of developing the diffusion model for local planning using the data collected to address the \textit{Myopic Observation} problem. The comprehensive workflow of our research is illustrated in Fig.~\ref{fig:overview}.

\subsection{Expert Policy Data}
In recent times, the swift advancement of reinforcement learning has unveiled novel solutions for robotic motion planning challenges. We employ the SAC algorithm to learn sophisticated navigation policies. This robot navigation problem is conceptualized as a Markov Decision Process. The expert policy employs the same structure as that in~\cite{yu2023pathrl}, which has been proven to be effective.%~\Jie{We can say: Use the same network architecture which is proved effective in [cite]}~\ynote{Done}
The policy's state space is bifurcated into (1) egocentric costmaps with dimensions of 84$\times$84, produced by a 3D laser sensor, encompassing a complete 360-degree view; (2) the relative target pose. The action space is bidimensional and continuous, denoting the linear velocity and the angle of the front wheel for Ackermann steering robots. Given that the linear velocity can assume negative values, this expert navigation policy accommodates backward movement. Building upon the meticulously crafted state and action spaces, and enabling the robot to swiftly reach its target without any collisions, the reward function for the reinforcement learning policy has been thoughtfully devised as follows:
\begin{equation}
    r_t = r_t^\textit{goal} + r_t^\textit{safe} + r_t^\textit{shaping} + r_t^\textit{back}, \label{eq:reward}
\end{equation}
where $r_t$ is the sum of four parts, $r_t^\textit{goal}$, $r_t^\textit{safe}$, $r_t^\textit{shaping}$ and $r_t^\textit{back}$.

Specifically, $r_t^\textit{goal}$ represents the reward awarded to the robot upon successfully reaching the designated local target:
\begin{equation}
r_t^{\textit{goal}} = \begin{cases}
r_{arr}, & \text{if target is reached,} \\
0, & \text{otherwise.} 
\end{cases}
\label{eq:goal}
\end{equation}

$r_t^{\textit{safe}}$ denotes the penalty applied to the robot in the event of a collision:
\begin{equation}
r_t^\textit{safe} = \begin{cases}
r_{col}, & \text{if collision,} \\
0, & \text{otherwise.} 
\end{cases}
\label{eq:safe}
\end{equation}

$r_t^\textit{shaping}$ denotes the reward-shaping mechanism that converts sparse rewards into dense rewards, thereby hastening the training process of reinforcement learning algorithms. The underlying design philosophy encompasses two main elements: (1) Actions diverting the robot from its local goal incur specific penalties; (2) To deter the agent from engaging in clever but impractical strategies—like spiraling near the target to gain higher rewards, counter to the necessity for swift navigation task completion—an additional fixed-value penalty is levied on each action.
\begin{equation}
r_t^\textit{shaping} = \varepsilon( \left\|p_{t-1} - p_g\right\|_2 - \left\|p_{t} - p_g\right\|_2 ) - r_{step},
\label{eq:shaping}
\end{equation}
where $p_{t}$ is the position of the robot at time $t$, $p_g$ is the position of the target point, $r_{step}$ is a fixed-value penalty, and $\varepsilon$ is a hyper-parameter.

The term $r_t^\textit{back}$ denotes a fixed-value penalty applied when the robot executes a reverse maneuver. This mechanism is designed to prompt the robot to reverse only when needed, instead of persistently backing up from start to end, thus aligning more closely with practical applications.
\begin{equation}
r_t^\textit{back} = \begin{cases}
r_{reverse}, & \text{if $v<0$,} \\
0, & \text{otherwise,} 
\end{cases}
\label{eq:back}
\end{equation}
where $r_{reverse}$ is also a fixed-value penalty and $v$ is robot's linear velocity at time $t$.
In our experiments, we set $r_{arr} = 500$, $r_{col} = -500$, $\varepsilon = 200$, $r_{step} = 5$ and $r_{reverse} = -10$.

\begin{algorithm}
\caption{Expert Data Collecting}
\textit{\# Collecting data on two types of preferred expert policies in each scenario}
\begin{algorithmic}[1]
\State \textit{Initialize the collection of expert data $\mathcal{T}^{train}$}
\For{$\pi_s \in \Pi$}
    \State Initialize episodes of collected data $N$, global path planning algorithm $A^*$ and size of sliding window $w$
    \State Clear the trajectory buffer $\mathcal{T}$
    \For{episode = $1 \  \textbf{to} \  N/2$}
        \For{step $t = 1 \  \textbf{to} \  T_{ep}$}
            \State $gp_t = A^*(p_t, g_t)$
            \State $a_t = \pi_s(s_t), s_t = \{ c_t, g_t \}$
            \State $\mathcal{T} \leftarrow \mathcal{T} \cup \{c_t, g_t, gp_t, a_t\}$
        \EndFor
    \EndFor
    \For{episode = $N/2 + 1 \  \textbf{to} \  N$}
        \For{step $t = 1 \  \textbf{to} \  T_{ep}$}
            \State $gp_t = A^*(p_t, g_t)$
            \State $lg_t = gp_t[w]$
            \State $a_t = \pi_s(s_t), s_t = \{ c_t, lg_t \}$
            \State $\mathcal{T} \leftarrow \mathcal{T} \cup \{c_t, g_t, gp_t, a_t\}$
        \EndFor
    \EndFor
    \State $\mathcal{T}^{train} \leftarrow \mathcal{T}^{train} \cup \mathcal{T}$
\EndFor
\end{algorithmic}
\end{algorithm}

Within the scope of our research, for every experimental scenario delineated in the paper, we meticulously trained an expert policy, subjecting each to an extensive training regimen spanning three million steps. Leveraging these expert policies, we collated datasets reflecting two strategic preferences: (1) The original expert policy data. While the reward-shaping mechanism outlined in Eq.~\eqref{eq:shaping} offers significant benefits, such as expedited training, diminished complexity, and the facilitation of rapid task completion, it tends to engender policies that overly greedy, lacking in long-term planning, and unsuitable for tasks in maze-like scenarios. (2) Expert policy data guided by global path planning. We utilized the A* algorithm to search for global paths and provided local targets to the expert policies through a sliding window approach. This type of policy overcomes the aforementioned issues but lacks efficiency in task completion, leading to detours in certain scenarios. Our objective is to enhance the policy model's performance by learning from the mixed data of these two types of preferred policies.

% \begin{algorithm}
% \caption{Expert Policy Training}
% \textit{\# Train a corresponding expert policy for each scenario}
% \begin{algorithmic}[1] % The [1] ensures line numbers are shown
% \State Initialize the collection of expert policies $\Pi$
% \For{scene = $1 \  \textbf{to} \  S$}
%     \State Initialize the policy network $\pi_s$, the value network, and various hyperparameters
%     \State Clear the experience buffer $\mathcal{D}$
%     \For{episode = $1 \  \textbf{to} \  E$}
%         \For{step $t = 1 \  \textbf{to} \  T_{ep}$}
%             \State $a_t = \pi(s_t), s_t = \{ c_t, g_t \}$
%             \State $s_{t+1}, r_t = step(a_t)$ % Corrected 'step(a)' to 'step(a_t)' for clarity
%             \State $\mathcal{D} \leftarrow \mathcal{D} \cup \{s_t, a_t, r_t, s_{t+1}\}$
%             \State Sample batch $\mathcal{B} \sim \mathcal{D}$ to update SAC network
%             \If{robot has stopped}
%                 \State $s_t = reset()$
%             \EndIf
%         \EndFor
%     \EndFor
%     \State $\Pi \leftarrow \Pi \cup \pi_s$
% \EndFor
% \end{algorithmic}
% \end{algorithm}

\begin{algorithm}
\caption{LDP Training and Evaluation}
\textit{\# LDP Training}
\begin{algorithmic}[1]
\State initialize training iterations $N$, batch size $M$
\For{$\mathcal{T} \subset \mathcal{T}^{train}$}
    \For{$n = 1 \  \textbf{to} \  N$}
        \State $\mathcal{B} \sim \{C_i, G_i, GP_i, A^0_i\}_{i=1}^M$
        \State randomly select a diffusion timestep $k \sim \mathcal{U}(1, K)$ and acquire noisy action sequences $A^k$
        \State compute loss function $\mathcal{L}(\theta)$ and update LDP
    \EndFor
\EndFor
\end{algorithmic}
\textit{\# LDP Evaluation}
\begin{algorithmic}[1]
\State initialize sample scale $\beta$ and classifier-free guidance scale $\omega$
\For{episode = $1 \  \textbf{to} \  E$}
    \For{step $t = 1 \  \textbf{to} \  T_{ep}$}
        \State sample $A^K \sim \mathcal{N}(0, \beta I)$
        \For{$k = K \  \textbf{to} \  1$}
            \State $\bar{\epsilon}=\epsilon_\theta(A^k, k) + \omega (\epsilon_\theta(A^k, \mathcal{O}, k) - \epsilon_\theta(A^k, k))$
            \State $(\mu_{k-1}, \sum_{k-1}) \leftarrow \text{Denoise}(A^k, \bar{\epsilon})$
            \State $A^{k-1} \sim \mathcal{N}(\mu_{k-1}, \beta \sum_{k-1})$
        \EndFor
        \State generate the final action sequence $A^0$
        \State execute each action within the $A^0$ sequence, requiring $T_a$ time steps
        \State obtain the next robot observation for the subsequent decision-making
    \EndFor
\EndFor
\end{algorithmic}
\end{algorithm}

\subsection{Local Diffusion Planner}
The objective of our research is to develop a local motion planning algorithm for robots, by leveraging multimodal expert strategy data encompassing a variety of environments and preferences. Therefore, we formulate the task as a conditional generation problem via diffusion model:
\[
    max_\theta \mathbb{E}_{\tau \sim \mathcal{T}, \tau = \{\mathcal{O}, A\}} \log p_\theta(A^0|\mathcal{O}),
\]
where $\tau$ is the expert trajectory data used for training, $A^0$ is the final generated sequence of actions, $\mathcal{O}$ represents the robot's observations serving as conditions for the diffusion model, and $p_\theta$ refers to the reverse denoising process within the diffusion model.

In this paper, we structure the training data according to a receding-horizon action prediction framework, as outlined in~\cite{chi2023diffusion}. Here, \(\tau_t = \{ \mathcal{O}_t, A_t \}\) signifies the chosen robot training trajectory at time \(t\), where \(\mathcal{O}_t\) and \(A_t\) respectively denote the corresponding observation sequence \(\{ o_{t-(T_o-1)}, \ldots, o_{t-1}, o_t \}\) and action sequence \(\{ a_t, a_{t+1}, \ldots, a_{t+(T_a-1)} \}\) for that trajectory. \(T_o\) and \(T_a\) represent the lengths of the observation and action sequences, respectively, with the length of the trajectory \(\tau\), denoted as \(T_{\tau}\), being equal to \(T_o + T_a - 1\). It should be emphasized that, throughout this paper, all superscripts associated with time instances refer to diffusion time steps, and all subscripts associated with time instances pertain to motion time steps.

As we have discussed earlier, $\mathcal{O}$ consists of three parts: costmaps $C$, goals $G$, and global paths $GP$, i.e., $\mathcal{O} = \{ C, G, GP \}$. It is important to note that here, the global paths act merely as conditions for the diffusion process, rather than providing an additional local target for the policy during inference, as is done in our method of collecting expert policy data.

The training of the model leverages the DDPM algorithm, incorporating classifier-free guidance~\cite{ho2022classifier} for its execution. The ultimate action sequence, $A^0$, is derived by initially sampling from Gaussian noise $A^K$. Through adjacent diffusion time steps, from $A^k$ to $A^{k-1}$, the action sequence is subjected to noise perturbation. This methodical alteration facilitates the denoising and refinement of the sequence itself. The perturbation noise $\bar{\epsilon}_\theta(A^k, \mathcal{O}, k)$ is defined as follows~\cite{nichol2021glide}:
\begin{equation}
    \epsilon_\theta(A^k, k) + \omega (\epsilon_\theta(A^k, \mathcal{O}, k) - \epsilon_\theta(A^k, k)),
    \label{eq:perturb}
\end{equation}
where $\omega > 1$ represents the guidance scale, a factor in identifying the expert action sequence that optimally aligns with the robot's current observations from the expert dataset, and $\epsilon_\theta$ denotes the noise model. 
Eq.~\eqref{eq:perturb} is inspired by an implicit classifier $p(\mathcal{O}|A^k) \propto p(A^k|\mathcal{O}) / p(A^k)$. The gradient of the logarithmic probability of this classifier $\nabla_{A^k} \log p(\mathbf{O}|A^k) \propto \nabla_{A^k} \log p(A^k|\mathcal{O}) - \nabla_{A^k} \log p(A^k) \propto \epsilon(A^k, \mathcal{O}, k) - \epsilon(A^k, k)$ is utilized to guide the generation of $\bar{\epsilon}$.
In the training phase, we aim to optimize the reverse diffusion process $p_\theta$, which is parameterized by the noise model, pursuing the following objective:
\begin{equation}
\mathcal{L}(\theta) := \mathbb{E}_{k \sim \mathcal{U}(1, K), \epsilon \sim \mathcal{N}(0, I)}[\left\| \epsilon - \epsilon_\theta(A^k, \mathcal{O}, k) \right\|^2].
\label{eq:loss}
\end{equation}

In the inference phase, the ultimate expert action sequence is derived via a stepwise sampling process, as outlined in the formula $A^{k-1} \sim \mathcal{N}(\mu_\theta(A^{k-1}, \mathcal{O}, k-1), \beta \sum_{k-1})$.

The design of the diffusion model network structure follows the approach of work~\cite{chi2023diffusion}, which has been validated as efficient and outstanding.

\section{Experiments}
\subsection{Environments, baselines, and metrics}

\begin{figure}[t]
    \centering
    \subfigure[static]{\includegraphics[width=0.24\linewidth]{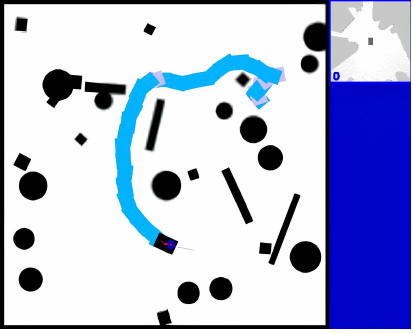}}
    \subfigure[dynamic]{\includegraphics[width=0.24\linewidth]{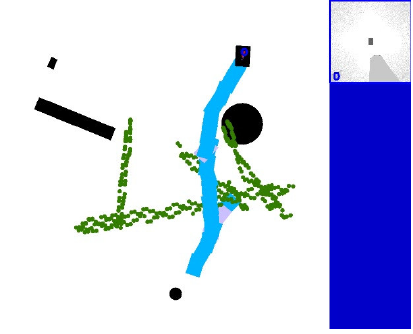}}
    \subfigure[maze]{\includegraphics[width=0.24\linewidth]{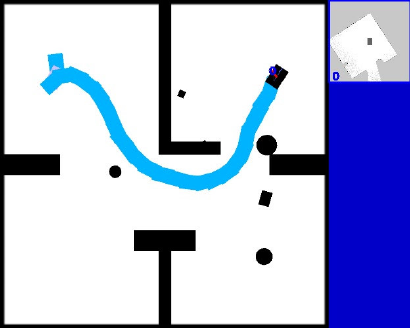}}
    \subfigure[zigzag]{\includegraphics[width=0.24\linewidth]{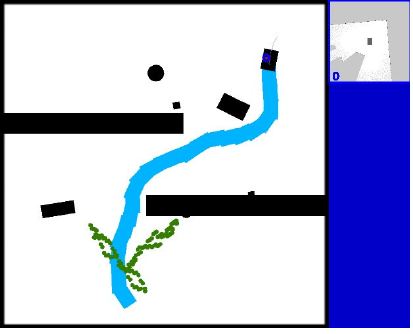}}
    \caption{Four different simulation scenarios are displayed. The black rectangles and circles are obstacles, the green dots represent pedestrian trajectories, and the blue box on the right shows the robot's local sensor map.}
    \label{fig:scene}
\end{figure}

In this study, we meticulously gather expert data reflecting two preferences across three diverse scenarios, subsequently training corresponding navigation policies. Besides, we evaluate the performance of models trained with mixed policy data and assess their zero-shot generalization capabilities in unseen scenarios. Fig.~\ref{fig:scene} illustrates environments for four different scenarios respectively, where a, c, and d represent enclosed scenarios, while b represents an open scenario.

We compare \texttt{LDP} with the following baseline algorithms: LSTM-GMM~\cite{mandlekar2021matters}, IBC~\cite{florence2022implicit}, and DT~\cite{chen2021decision}. We strive to adjust the training details to optimize the performance of the methods as well as possible. Notice that, the input content for these baseline algorithms consists of $\{C, G, GP\}$, which aligns with the guiding conditions of \texttt{LDP}.
% \begin{itemize}
% \item \textbf{GMM} belongs to the Explicit Policy in imitation learning, using the MDNs method to combine Categorical and Gaussian distributions to represent multimodal distributions.
% \item \textbf{IBC} belongs to the Implicit Policy in imitation learning, using the argmin operation and a continuous energy function to represent the policy instead of an explicit function mapping.
% % \item \textbf{CQL}, an offline reinforcement learning method, introduces an additional regularization component to the standard Q-function learning objective. This is designed to mitigate the overestimation of Q-values for actions absent from the dataset, embodying a conservative approach to policy updates. CQL~\cite{kumar2020conservative}
% \item \textbf{DT}, an offline reinforcement learning method, views the reinforcement learning task as a sequence modeling task, using the Transformer architecture to learn policies by directly optimizing cumulative rewards in an end-to-end manner.  
% Essentially, it is a form of Reward-based imitation learning.
% \item \textbf{Diffuser} employs diffusion models (DDPM) with classifier guidance for the training of agent policies.
% \end{itemize}

We introduce five distinct metrics to systematically assess the navigation performance of all methods within a range of scenarios.
\begin{itemize}
    \item \textit{Success rate (SUCC)}: the rate of episodes in which the robot reaches the target pose without collision.
    \item \textit{Collision rate (Coll)}: the rate of episodes in which the robot collides.
    \item \textit{Stuck rate (Stuck)}: the rate of episodes in which the robot stucks.
    \item \textit{Average time (TIME)}: the average cost time at each success episode.
    \item \textit{Success weighted by Path Length (SPL)}~\cite{yokoyama2021success}: assess navigation performance by examining both the success rate and the length of the robot's trajectory.
\end{itemize}

\begin{table}[htbp]
\centering
\caption{Hyperparameters}
\label{tab:hyper}
\renewcommand{\arraystretch}{1.0} % Adjusts the row height
\resizebox{\columnwidth}{!}{
\begin{tabular}{>{\centering\arraybackslash}p{5.0cm} 
                >{\centering\arraybackslash}p{2.0cm} 
                }
\toprule
\textbf{Hyperparameter} & \textbf{Value} \\
\midrule
\rowcolor[gray]{.9} \multicolumn{2}{c}{Hyperparameters for expert policy} \\
Training steps & $3\times 10^6$ \\
Encoder\&Actor\&Critic learning rate & $5\times 10^{-4}$ \\
Temperature learning rate & $10^{-4}$ \\
Init temperature & $0.01$ \\
Batch size & $1024$ \\
Obstacle collision radius for A* & $0.5$ \\
Size of the sliding window & $20$ \\
\rowcolor[gray]{.9} \multicolumn{2}{c}{Hyperparameters for local diffusion planner} \\
Training epochs & 300 \\
Episodes of each scenario expert data & 2000 \\
Batch Size & 1024 \\
Learning rate & $10^{-4}$ \\
Weight decay & $10^{-6}$ \\
Prediction horizon $T$ & $8$ \\
Observation horizon $T_o$ & $2$ \\
Action horizon $T_a$ & $4$ \\
Number of diffusion iterations & $100$ \\

\bottomrule
\end{tabular}}
\end{table}

\subsection{Experiments on simulation scenarios}
In the following, the experimental results for various methods are presented based on the average outcomes from 1,000 randomly generated environments for each scenario. The training data for each scenario consists of 2,000 episodes. % All algorithms in the paper undergo training for an equal number of epochs.

\begin{table*}[h]
\centering
\caption{Performance of Different Methods}
\label{tab:comparative}
\renewcommand{\arraystretch}{1.0} % Adjusts the row height
% \resizebox{2.0\columnwidth}{!}{
\begin{tabular}{>{\centering\arraybackslash}p{1.6cm} 
                >{\centering\arraybackslash}p{2.2cm} 
                >{\centering\arraybackslash}p{2.2cm} 
                >{\centering\arraybackslash}p{2.2cm} 
                >{\centering\arraybackslash}p{2.8cm}
                >{\centering\arraybackslash}p{2.2cm}
                }
\toprule
\textbf{Methods} & \textbf{Succ(\%)}$\uparrow$ & \textbf{Coll(\%)}$\downarrow$ & \textbf{Stuck(\%)}$\downarrow$ & \textbf{Time(t)}$\downarrow$ & \textbf{SPL}$\uparrow$ \\
\midrule
\rowcolor[gray]{.9} \multicolumn{6}{c}{Static Scenario (28 Obstacles)} \\
LSTM-GMM & 0.475-0.427-0.639 & 0.206-0.110-0.201 & 0.319-0.463-0.161 & 44.797-42.964-39.113 & 0.209-0.202-0.310\\
IBC & 0.635-0.829-0.741 & 0.164-0.127-0.073 & 0.201-0.044-0.186 & 41.612-31.971-37.109 & 0.299-0.464-0.404\\
% CQL & 02 & 30 & 11.22 & 98 & 22\\
DT & 0.679-0.772-0.752 & 0.134-0.135-0.110 & 0.187-0.093-0.138 & 36.891-30.092-36.242 & 0.347-0.441-0.379\\
% Diffuser & 02 & 30 & 11.22 & 98 & 22\\
LDP & \textbf{0.952}-\textbf{0.955}-\textbf{0.957} & \textbf{0.020}-\textbf{0.014}-\textbf{0.014} & \textbf{0.028}-\textbf{0.031}-\textbf{0.029} & \textbf{24.365}-\textbf{24.146}-\textbf{23.636} & \textbf{0.648}-\textbf{0.649}-\textbf{0.654}\\
\rowcolor[gray]{.9} \multicolumn{6}{c}{Dynamic Scenario (4 Obstacles \& 4 Pedestrians)} \\
LSTM-GMM & 0.693-0.715-0.657 & 0.265-0.251-0.270 & 0.042-0.034-0.066 & 31.543-28.012-34.716 & 0.396-0.443-0.338\\
IBC & 0.422-0.719-0.732 & 0.451-0.253-\textbf{0.227} & 0.127-0.028-0.041 & 52.150-34.152-32.657 & 0.191-0.369-0.415\\
% CQL & 02 & 30 & 11.22 & 98 & 22\\
DT & 0.508-0.744-0.708 & 0.487-0.243-0.258 & \textbf{0.005}-0.013-0.034 & 20.738-\textbf{20.980}-23.045 & 0.375-\textbf{0.533}-0.481\\
% Diffuser & 02 & 30 & 11.22 & 98 & 22\\
LDP & \textbf{0.740}-\textbf{0.748}-\textbf{0.755} & \textbf{0.255}-\textbf{0.242}-0.238 & \textbf{0.005}-\textbf{0.010}-\textbf{0.007} & \textbf{20.480}-22.272-\textbf{20.943} & \textbf{0.533}-0.523-\textbf{0.536}\\
\rowcolor[gray]{.9} \multicolumn{6}{c}{Maze-like Scenario (Maze Map \& 6 Obstacles)} \\
LSTM-GMM & 0.418-0.796-0.746 & 0.459-0.152-0.198 & 0.123-0.052-0.056 & 52.189-32.517-33.521 & 0.190-0.407-0.371\\
IBC & 0.715-0.812-0.777 & 0.221-0.144-0.114 & 0.064-0.044-0.109 & 25.692-22.052-30.773 & 0.427-0.507-0.426\\
% CQL & 02 & 30 & 11.22 & 98 & 22\\
DT & 0.686-0.861-0.785 & 0.237-0.106-0.178 & 0.077-0.033-0.037 & 27.657-20.092-30.231 & 0.395-0.564-0.421\\
% Diffuser & 02 & 30 & 11.22 & 98 & 22\\
LDP & \textbf{0.915}-\textbf{0.930}-\textbf{0.930} & \textbf{0.065}-\textbf{0.057}-\textbf{0.058} & \textbf{0.020}-\textbf{0.013}-\textbf{0.012} & \textbf{20.248}-\textbf{19.261}-\textbf{19.163} & \textbf{0.609}-\textbf{0.637}-\textbf{0.631}\\
\rowcolor[gray]{.9} \multicolumn{6}{c}{Zigzag Scenario (Unseen)} \\
LSTM-GMM & 0.500 & 0.439 & 0.061 & 38.002 & 0.271\\
IBC & 0.532 & 0.330 & 0.138 & 33.490 & 0.333\\
% CQL & 02 & 30 & 11.22 & 98 & 22\\
DT & 0.464 & 0.451 & 0.085 & 32.464 & 0.283\\
% Diffuser & 02 & 30 & 11.22 & 98 & 22\\
LDP & \textbf{0.720} & \textbf{0.267} & \textbf{0.013} & \textbf{22.264} & \textbf{0.510}\\
\bottomrule
\end{tabular}
\end{table*}

\subsubsection{Comparative experiments}
Tab.~\ref{tab:comparative} presents the navigation performance of all algorithms outlined in our paper. Each entry comprises three values: the performance metrics of models trained on single-scenario data (2000 episodes), the performance metrics of models trained on single-scenario data (6000 episodes), and the performance metrics of models trained on mixed data from three scenarios (6000 episodes). The \texttt{LDP} algorithm surpasses other baseline algorithms in success rate, runtime, and SPL. Comparing the first and last two items, we can conclude that increasing the quantity of training data to some extent can enhance the navigation performance across a wide range of algorithms and scenarios. Contrasting the second and third items, we can infer that enriching the diversity of data while maintaining the same training data quantity poses a challenge for baseline algorithms, resulting in a decline in navigation performance. However, thanks to the superior distribution modeling capability of \texttt{LDP}, its navigation performance may remain stable or even improve.
An interesting observation is that LSTM-GMM, when trained on data from a single dense static scenario, cannot accurately reach the target point. It tends to wander near the target point, resulting in a high stuck rate. Even increasing the training data volume does not resolve this issue. However, introducing data from diverse scenarios can significantly enhance the performance of LSTM-GMM.
Notably, in the zigzag scenario, the performance of \texttt{LDP} underscores its robust zero-shot generalization capability in unseen scenarios.

\subsubsection{Ablation study}
\begin{figure}[h]
    \centering
    \includegraphics[width=0.8\columnwidth, height=0.15\textwidth]{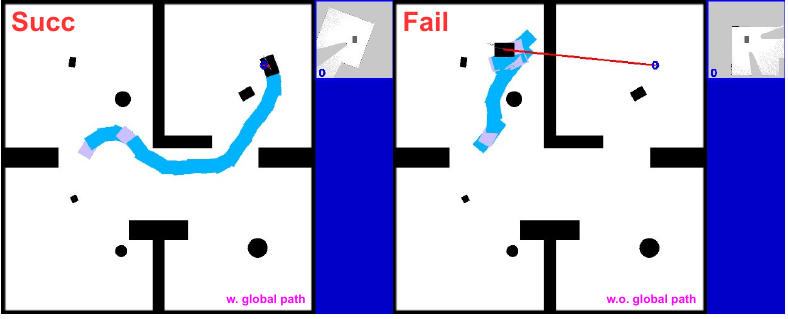}
    \caption{Global Path Influence: Navigation Success vs. Failure in One Scene}
    \label{fig:gp}
\end{figure}

In the design of the \texttt{LDP} approach, we have additionally introduced global paths $GP$ as conditions for the diffusion model to guide wiser decision-making in complex scenarios. Tab.~\ref{tab:gpath} illustrates that \texttt{LDP} outperforms \texttt{LDP} without $GP$, particularly in dense static and maze-like scenarios. In Fig.~\ref{fig:gp}, we showcase a maze-like scenario where \texttt{LDP} effectively navigates around maze walls and reaches the target point, while \texttt{LDP} without $GP$ gets obstructed by the walls, failing to complete the navigation task. The experimental results suggest that additional $GP$ conditions can better assist \texttt{LDP} in modeling data distributions and guiding wiser decision-making.

\begin{table}[htbp]
\centering
\caption{Performance Guided by Global Path Conditions}
\label{tab:gpath}
\renewcommand{\arraystretch}{1.0} % Adjusts the row height
\resizebox{\columnwidth}{!}{
\begin{tabular}{>{\centering\arraybackslash}p{1.8cm} 
                >{\centering\arraybackslash}p{1.0cm} 
                >{\centering\arraybackslash}p{1.0cm} 
                >{\centering\arraybackslash}p{1.1cm} 
                >{\centering\arraybackslash}p{1.0cm}
                >{\centering\arraybackslash}p{0.8cm}
                }
\toprule
\textbf{Methods} & \textbf{Succ(\%)}$\uparrow$ & \textbf{Coll(\%)}$\downarrow$ & \textbf{Stuck(\%)}$\downarrow$ & \textbf{Time(t)}$\downarrow$ & \textbf{SPL}$\uparrow$ \\
\midrule
\rowcolor[gray]{.9} \multicolumn{6}{c}{Static Scenario (28 Obstacles)} \\
LDP & \textbf{0.952} & 0.020 & \textbf{0.028} & \textbf{24.365} & \textbf{0.648}\\
LDP $w.o.\ GP$ & 0.934 & \textbf{0.017} & 0.049 & 25.264 & 0.632\\
\rowcolor[gray]{.9} \multicolumn{6}{c}{Dynamic Scenario (4 Obstacles \& 4 Pedestrians)} \\
LDP & \textbf{0.740} & 0.255 & \textbf{0.005} & \textbf{20.480} & \textbf{0.533}\\
LDP $w.o.\ GP$ & 0.737 & \textbf{0.248} & 0.015 & 21.903 & 0.521\\
\rowcolor[gray]{.9} \multicolumn{6}{c}{Maze-like Scenario (Maze Map \& 6 Obstacles)} \\
LDP & \textbf{0.915} & \textbf{0.065} & \textbf{0.020} & 20.248 & \textbf{0.609}\\
LDP $w.o.\ GP$ & 0.877 & 0.095 & 0.028 & \textbf{19.689} & 0.595\\   
\bottomrule
\end{tabular}}
\end{table}

\begin{figure}[h]
    \centering
    \includegraphics[width=\columnwidth]{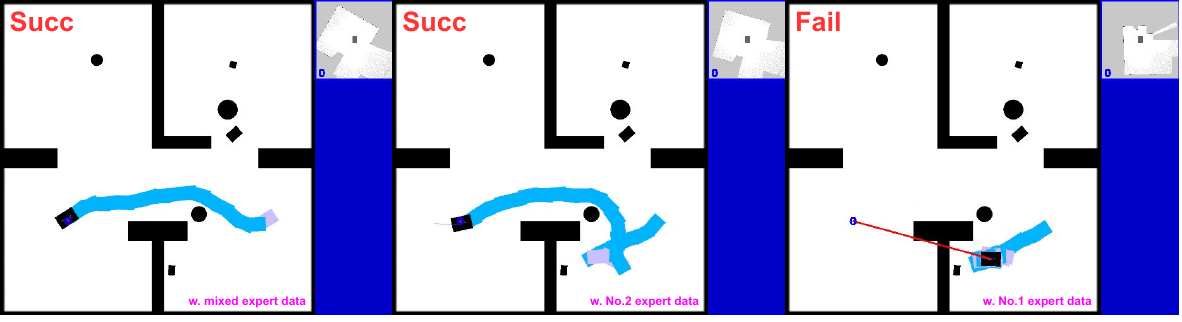}
    \caption{Performance of policies learned under expert data with different preferences.}
    \label{fig:pf}
\end{figure}

In our paper, we collected expert data reflecting two different preferences. Therefore, in Tab.~\ref{tab:prefer}, we explore the impact of training data composition with varying preferences on model performance. All experiments are trained on the same amount of data for the same training steps.
%~\Jie{Emphasize: Training All Methods with the same training steps}~\ynote{Done} 
The ``No.1 expert'', ``No.2 expert'', and ``mixed expert'' respectively represent data collected from original SAC (2000 episodes), SAC guided by global paths (2000 episodes), and a mixture of both, with 1000 episodes each. Our experimental results demonstrate that mixed preference data can enhance the policy's performance in dynamic and maze-like scenarios. While in static scenes, the success rate of the mixed data policy slightly lags behind that of the No.2 expert data policy, it outperforms in terms of average execution time and SPL. In the maze-like scenario presented in Fig.~\ref{fig:pf}, \texttt{LDP} with mixed expert data efficiently completes the navigation task. Conversely, \texttt{LDP} with No.2 expert data initially encounters obstacles due to erroneous decisions but eventually succeeds after prolonged exploration, while \texttt{LDP} with No.1 expert data remains trapped and unable to finish the task.
Providing data with mixed preferences is meaningful as it allows the policy to leverage the advantages of learning different preference policies, resulting in more efficient and accurate task completion.
%~\Jie{The conclusion proves the importance of global information and the Figure gives the case that lack of GP will make agent go to the local optimal point.}~\ynote{Done}

\begin{table}[htbp]
\centering
\caption{Performance Trained by Expert Data with Two Preferences}
\label{tab:prefer}
\renewcommand{\arraystretch}{1.0} % Adjusts the row height
\resizebox{\columnwidth}{!}{
\begin{tabular}{>{\raggedright\arraybackslash}p{2.5cm} 
                >{\centering\arraybackslash}p{1.0cm} 
                >{\centering\arraybackslash}p{0.9cm} 
                >{\centering\arraybackslash}p{1.1cm}
                >{\centering\arraybackslash}p{0.9cm}
                >{\centering\arraybackslash}p{0.5cm}
                }
\toprule
\textbf{Methods} & \textbf{Succ(\%)}$\uparrow$ & \textbf{Coll(\%)}$\downarrow$ & \textbf{Stuck(\%)}$\downarrow$ & \textbf{Time(t)}$\downarrow$ & \textbf{SPL}$\uparrow$ \\
\midrule
\rowcolor[gray]{.9} \multicolumn{6}{c}{Static Scenario (28 Obstacles)} \\
LDP $w.$ No.1 expert& 0.908 & 0.027 & 0.065 & \textbf{23.667} & 0.633\\
LDP $w.$ No.2 expert & \textbf{0.956} & \textbf{0.012} & 0.032 & 24.877 & 0.642\\
LDP $w.$ mixed expert & 0.952 & 0.020 & \textbf{0.028} & 24.365 & \textbf{0.648}\\
\rowcolor[gray]{.9} \multicolumn{6}{c}{Dynamic Scenario (4 Obstacles \& 4 Pedestrians)} \\
LDP $w.$ No.1 expert & 0.732 & \textbf{0.248} & 0.020 & 20.499 & 0.530\\
LDP $w.$ No.2 expert & 0.726 & 0.266 & 0.008 & 22.777 & 0.494\\
LDP $w.$ mixed expert & \textbf{0.740} & 0.255 & \textbf{0.005} & \textbf{20.480} & \textbf{0.533}\\
\rowcolor[gray]{.9} \multicolumn{6}{c}{Maze-like Scenario (Maze Map \& 6 Obstacles)} \\
LDP $w.$ No.1 expert & 0.868 & 0.099 & 0.033 & \textbf{19.800} & 0.589\\
LDP $w.$ No.2 expert & 0.873 & 0.094 & 0.033 & 19.904 & 0.592\\
LDP $w.$ mixed expert & \textbf{0.915} & \textbf{0.065} & \textbf{0.020} & 20.248 & \textbf{0.609}\\
\bottomrule
\end{tabular}}
\end{table}

In the field of robot navigation, global paths can provide global information to local planners to assist them in making better decisions. Similarly, global maps $GM$ can serve this role as well. We conducted experimental comparisons of two diffusion model conditions in Tab.~\ref{tab:diffcond}. All experiments are trained for the same number of steps. \texttt{LDP} with $GP$ demonstrates higher success rates and SPL with slightly longer average running times in dense static and maze-like complex structured scenarios. In dynamic scenarios, \texttt{LDP} with $GP$ exhibits slightly lower success rates than \texttt{LDP} with $GM$, but it boasts shorter average running times and higher SPL values. The experimental results suggest that $GP$ offers more direct and efficient guidance compared to $GM$, particularly excelling in complex structured scenarios such as dense static and maze-like environments.
%~\Jie{Training with the same training steps and go to the conclusion: Diverse data is matter!!}~\ynote{Done}

\begin{table}[htbp]
\centering
\caption{Performance Guided by Different Global Conditions}
\label{tab:diffcond}
\renewcommand{\arraystretch}{1.0} % Adjusts the row height
\resizebox{\columnwidth}{!}{
\begin{tabular}{>{\centering\arraybackslash}p{2.5cm} 
                >{\centering\arraybackslash}p{1.0cm} 
                >{\centering\arraybackslash}p{0.9cm} 
                >{\centering\arraybackslash}p{1.1cm}
                >{\centering\arraybackslash}p{0.9cm}
                >{\centering\arraybackslash}p{0.5cm}
                }
\toprule
\textbf{Methods} & \textbf{Succ(\%)}$\uparrow$ & \textbf{Coll(\%)}$\downarrow$ & \textbf{Stuck(\%)}$\downarrow$ & \textbf{Time(t)}$\downarrow$ & \textbf{SPL}$\uparrow$ \\
\midrule
\rowcolor[gray]{.9} \multicolumn{6}{c}{Static Scenario (28 Obstacles)} \\
LDP $w.\ GP$ & \textbf{0.952} & \textbf{0.020} & \textbf{0.028} & 24.365 & \textbf{0.648}\\
LDP $w.\ GM$ & 0.930 & 0.026 & 0.044 & \textbf{24.251} & 0.634\\
\rowcolor[gray]{.9} \multicolumn{6}{c}{Dynamic Scenario (4 Obstacles \& 4 Pedestrians)} \\
LDP $w.\ GP$ & 0.740 & 0.255 & \textbf{0.005} & \textbf{20.480} & \textbf{0.533}\\
LDP $w.\ GM$ & \textbf{0.744} & \textbf{0.238} & 0.018 & 21.777 & 0.527\\
\rowcolor[gray]{.9} \multicolumn{6}{c}{Maze-like Scenario (Maze Map \& 6 Obstacles)} \\
LDP $w.\ GP$ & \textbf{0.915} & \textbf{0.065} & \textbf{0.020} & 20.248 & \textbf{0.609}\\
LDP $w.\ GM$ & 0.903 & 0.077 & \textbf{0.020} & \textbf{19.596} & 0.608\\
\bottomrule
\end{tabular}}
\end{table}

\subsection{Deploy to real-world Ackermann steering robot}

We've implemented the \texttt{LDP} algorithm on a real Ackerman robot to evaluate its performance in real-world situations. This experimental robot is built on the Agilex hunter2.0 chassis and features a 32-line 3D RoboSense LiDAR. It's powered by an RTX 3090 GPU and measures $0.95m \times 0.75m \times 1.45m$ in size. For a more detailed overview of the simulation and physical experiment results, please refer to our video.
% To enhance the real-time performance of LDP on this physical robot, we've transitioned from DDPM to DDIM and reduced the number of steps during the inference process.

\begin{figure}[h]
    \centering
    \includegraphics[width=\columnwidth, height=0.4\columnwidth]{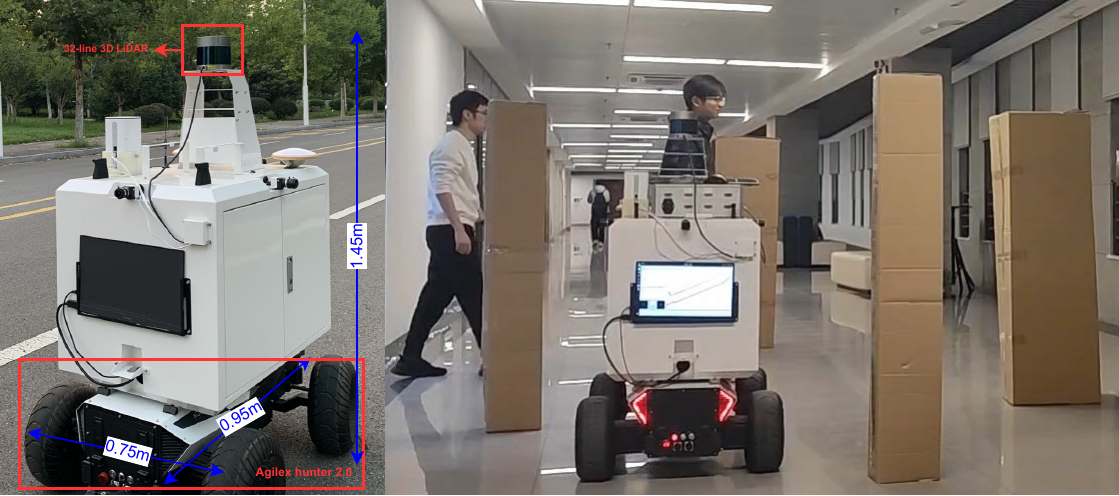}
    \caption{Schematic diagram of real robots and test scenarios.}
    \label{fig:real}
\end{figure}

\section{Conclusions and Future Work}
In this paper, we introduce a novel local diffusion planner, \texttt{LDP}, designed for robot collision avoidance, employing diffusion processes. Our approach involves gathering expert data representing two different preferences from three diverse scenarios to train our model. Our series of experiments indicate that \texttt{LDP} exhibits better navigation performance, and stronger robustness in learning from expert data across scenes. Additionally, it can learn wiser and more visionary policies from multi-preference expert data and demonstrate strong generalization ability in unseen scenarios. Our real-world experiments also demonstrate the practical value of \texttt{LDP}.

In future work, two aspects could be further explored: (1) Collecting higher quality and more diverse expert data to train superior navigation policies; (2) Improving the real-time performance of \texttt{LDP}. Using flow-based methods~\cite{liu2022flow} or consistency models~\cite{song2023consistency} instead of DDPM to accelerate diffusion model sampling would facilitate the practical deployment of the \texttt{LDP} method.

\bibliographystyle{IEEEtran}
\bibliography{LDiffPlanner}
\end{document}